\documentclass[11pt,a4paper]{article}
\usepackage[hyperref]{acl2018}
\usepackage{times}
\usepackage{latexsym}
\usepackage{array}

\usepackage{graphicx}
\usepackage{amsmath, amssymb}
\usepackage{svg}
\usepackage{multirow}
\usepackage[linesnumbered]{algorithm2e}
\aclfinalcopy 

\newcommand\emoji[1]
{$\vcenter{\hbox{\includegraphics[scale=0.31]{emoji/a#1.pdf}}}$}
\newcommand\bigmoji[1]
{$\vcenter{\hbox{\includegraphics[scale=0.33]{emoji/a#1.pdf}}}$}

\title{\textsc{MojiTalk}: Generating Emotional Responses at Scale}

\author{Xianda Zhou\\
Dept. of Computer Science and Technology\\
Tsinghua University\\
Beijing, 100084 China\\
  {\tt zhou-xd13@mails.tsinghua.edu.cn} \\\And
  William Yang Wang \\
  Department of Computer Science \\
  University of California, Santa Barbara\\
  Santa Barbara, CA 93106 USA\\
  {\tt william@cs.ucsb.edu} \\}

\date{} 

\begin{document}
\maketitle
\begin{abstract}
Generating emotional language is a key step towards building empathetic natural language processing agents. However, a major challenge for this line of research is the lack of large-scale labeled training data, and previous studies are limited to only small sets of human annotated sentiment labels. Additionally, explicitly controlling the emotion and sentiment of generated text is also difficult. In this paper, we take a more radical approach: we exploit the idea of leveraging Twitter data that are naturally labeled with emojis.

We collect a large corpus of Twitter conversations that include emojis in the response and assume the emojis convey the underlying emotions of the sentence. We investigate several conditional variational autoencoders training on these conversations, which allow us to use emojis to control the emotion of the generated text. Experimentally, we show in our quantitative and qualitative analyses that the proposed models can successfully generate high-quality abstractive conversation responses in accordance with designated emotions.
\end{abstract}

\section{Introduction}
A critical research problem for artificial intelligence is to design intelligent agents that can perceive and generate human emotions. In the past decade, there has been significant progress in sentiment analysis~\cite{pang2002thumbs,pang2008opinion,liu2012sentiment} and natural language understanding---e.g., classifying the sentiment of online reviews. To build empathetic conversational agents, machines must also have the ability to learn to generate emotional sentences.

One of the major challenges is the lack of large-scale, manually labeled emotional text datasets. Due to the cost and complexity of manual annotation, most prior research studies primarily focus on small-sized labeled datasets~\cite{pang2002thumbs,maas2011learning,socher2013recursive}, which are not ideal for training deep learning models with a large number of parameters.

In recent years, a handful of medium to large scale, emotional corpora in the area of emotion analysis~\cite{go2016sentiment140} and dialog~\cite{li2017dailydialog} are proposed. However, all of them are limited to a traditional, small set of labels, for example, ``happiness,'' ``sadness,'' ``anger,'' etc. or simply binary ``positive'' and ``negative.'' Such coarse-grained classification labels make it difficult to capture the nuances of human emotion.

\begin{figure}[t]
\centering
\includegraphics[width=\linewidth]{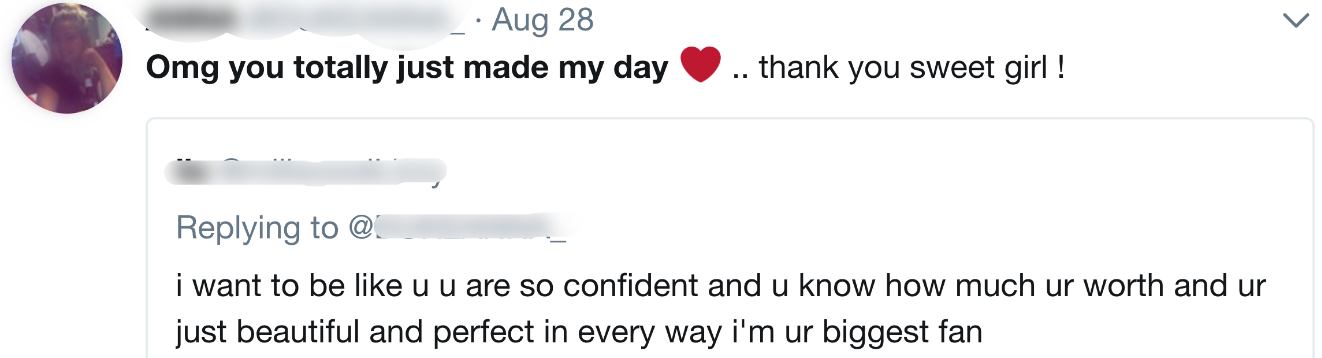}
\caption{An example Twitter conversation with emoji in the response (top). We collected a large amount of these conversations, and trained a reinforced conditional variational autoencoder model to automatically generate abstractive emotional responses given any emoji.}
\label{fig:example}
\end{figure}

To avoid the cost of human annotation, we propose the use of naturally-occurring emoji-rich Twitter data. We construct a dataset using Twitter conversations with emojis in the response. The fine-grained emojis chosen by the users in the response can be seen as the natural label for the emotion of the response. 

We assume that the emotions and nuances of emojis are established through the extensive usage by Twitter users. If we can create agents that are able to imitate Twitter users' language style when using those emojis, we claim that, to some extent, we have captured those emotions. Using a large collection of Twitter conversations, we then trained a conditional generative model to automatically generate the emotional responses.
Figure~\ref{fig:example} shows an example.

To generate emotional responses in dialogs, another technical challenge is to control the target emotion labels. In contrast to existing work~\cite{huang2017moodswipe} that uses information retrieval to generate emotional responses, the research question we are pursuing in this paper, is to design novel techniques that can generate abstractive responses of any given arbitrary emotions, without having human annotators to label a huge amount of training data.

To control the target emotion of the response, we investigate several encoder-decoder generation models, including a standard attention-based \textsc{seq2seq} model as the base model, and a more sophisticated CVAE model ~\cite{kingma2013auto,sohn2015learning}, as VAE is recently found convenient in dialog generation~\cite{zhao2017learning}.

To explicitly improve emotion expression, we then experiment with several extensions to the CVAE model, including a hybrid objective with policy gradient. The performance in emotion expression is automatically evaluated by a separate sentence-to-emoji classifier~\cite{felbo2017using}. Additionally, we conducted a human evaluation to assess the quality of the generated emotional text. 


Results suggest that our method is capable of generating state-of-the-art emotional text at scale. Our main contributions are three-fold:
\begin{itemize}
\item We provide a publicly available, large-scale dataset of Twitter conversation-pairs naturally labeled with fine-grained emojis.
\item We are the first to use naturally labeled emojis for conducting large-scale emotional response generation for dialog.
\item We apply several state-of-the-art generative models to train an emotional response generation system, and analysis confirms that our models deliver strong performance. 
\end{itemize}

In the next section, we outline related work on sentiment analysis and emoji on Twitter data, as well as neural generative models. Then, we will introduce our new emotional research dataset and formalize the task. Next, we will describe the neural models we applied for the task. Finally, we will show automatic evaluation and human evaluation results, and some generated examples. Experiment details can be found in supplementary materials.

\section{Related Work}

In natural language processing, sentiment analysis~\cite{pang2002thumbs} is an area that involves designing algorithms for understanding emotional text. 
Our work is aligned with some recent studies on using emoji-rich Twitter data for sentiment classification.
Eisner et al.~\shortcite{eisner2016emoji2vec} proposes a method for training emoji embedding \textsc{emoji2vec},
and combined with word2vec~\cite{mikolov2013efficient}, they apply the embeddings for sentiment classification.
DeepMoji~\cite{felbo2017using} is closely related to our study: It makes use of a large, naturally labeled Twitter emoji dataset, and train an attentive bi-directional long short-term memory network~\cite{hochreiter1997long} model for sentiment analysis. Instead of building a sentiment classifier, our work focuses on generating emotional responses, given the context and the target emoji.

Our work is also in line with the recent progress of the application of Variational Autoencoder (VAE)~\cite{kingma2013auto} in dialog generation. VAE~\cite{kingma2013auto} encodes data in a probability distribution, and then samples from the distribution to generate examples. However, the original frameworks do not support end-to-end generation. Conditional VAE (CVAE)~\cite{sohn2015learning,larsen2015autoencoding} was proposed to incorporate conditioning option in the generative process.
Recent research in dialog generation shows that language generated by VAE models enjoy significantly greater diversity than traditional \textsc{seq2seq} models~\cite{zhao2017learning}, which is a preferable property toward building a true-to-life dialog agents.


In dialog research, our work aligns with recent advances in sequence-to-sequence models~\cite{sutskever2014sequence} using long short-term memory networks~\cite{hochreiter1997long}. A slightly altered version of this model serves as our base model. Our modification enabled it to condition on single emojis. Li et al.~\shortcite{li2016deep} use a reinforcement learning algorithm to improve the vanilla sequence-to-sequence model for non-task-oriented dialog systems, but their reinforced and its follow-up adversarial models~\cite{li2017adversarial} also do not model emotions or conditional labels. 
Zhao et al.~\shortcite{zhao2017learning} recently introduced conditional VAE for dialog modeling, but neither did they model emotions in the conversations, nor explore reinforcement learning to improve results.
Given a dialog history, Xie et. al.'s work recommends suitable emojis for current conversation. Xie et. al.~\shortcite{xie2016neural}compress the dialog history to vector representation through a hierarchical RNN and then map it to a emoji by a classifier, while in our model, the representation for original tweet, combined with the emoji embedding, is used to generate a response.


\begin{table}[t]
\setlength\tabcolsep{2.1pt}
\renewcommand{\arraystretch}{1.6}
\small
\centering
\begin{tabular}{lr c lr c lr c lr}
\emoji{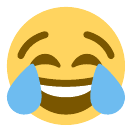}&184,500&~~~~&\emoji{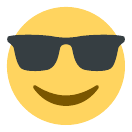}&9,505&~~~~&\emoji{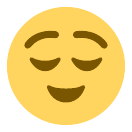}&5,558&~~~~&\emoji{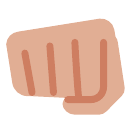}&2,771\\
\emoji{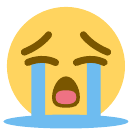}&38,479&~~~~&\emoji{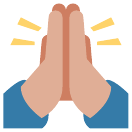}&9,455&~~~~&\emoji{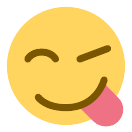}&5,114&~~~~&\emoji{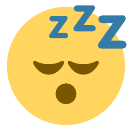}&2,532\\
\emoji{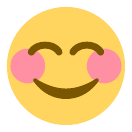}&30,447&~~~~&\emoji{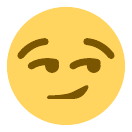}&9,298&~~~~&\emoji{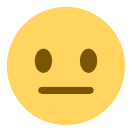}&5,026&~~~~&\emoji{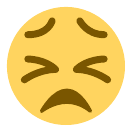}&2,332\\
\emoji{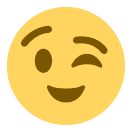}&25,018&~~~~&\emoji{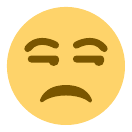}&8,385&~~~~&\emoji{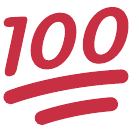}&4,738&~~~~&\emoji{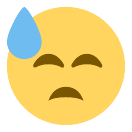}&2,293\\
\emoji{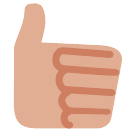}&19,832&~~~~&\emoji{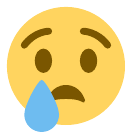}&8,341&~~~~&\emoji{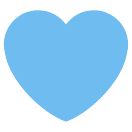}&4,623&~~~~&\emoji{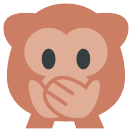}&1,698\\
\emoji{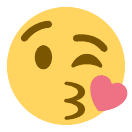}&16,934&~~~~&\emoji{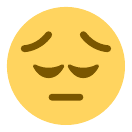}&8,293&~~~~&\emoji{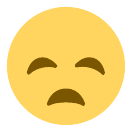}&4,531&~~~~&\emoji{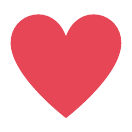}&1,534\\
\emoji{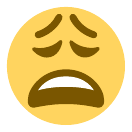}&17,009&~~~~&\emoji{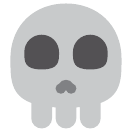}&8,144&~~~~&\emoji{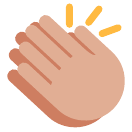}&4,287&~~~~&\emoji{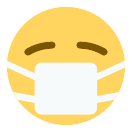}&1,403\\
\emoji{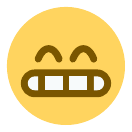}&15,563&~~~~&\emoji{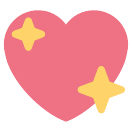}&7,101&~~~~&\emoji{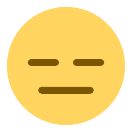}&4,205&~~~~&\emoji{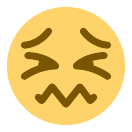}&1,258\\
\emoji{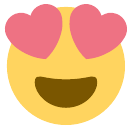}&15,046&~~~~&\emoji{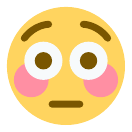}&6,939&~~~~&\emoji{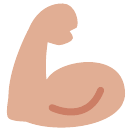}&4,066&~~~~&\emoji{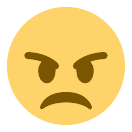}&1,091\\
\emoji{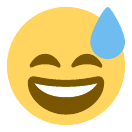}&14,121&~~~~&\emoji{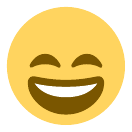}&6,769&~~~~&\emoji{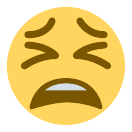}&3,973&~~~~&\emoji{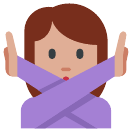}&698\\
\emoji{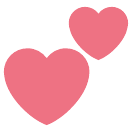}&13,887&~~~~&\emoji{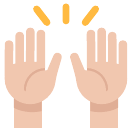}&6,625&~~~~&\emoji{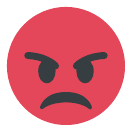}&3,841&~~~~&\emoji{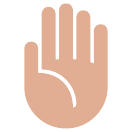}&627\\
\emoji{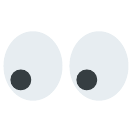}&13,741&~~~~&\emoji{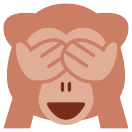}&6,558&~~~~&\emoji{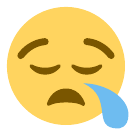}&3,863&~~~~&\emoji{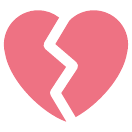}&423\\
\emoji{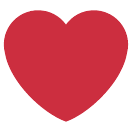}&13,147&~~~~&\emoji{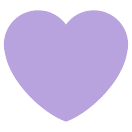}&6,374&~~~~&\emoji{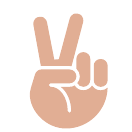}&3,236&~~~~&\emoji{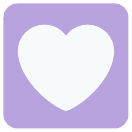}&250\\
\emoji{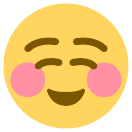}&10,927&~~~~&\emoji{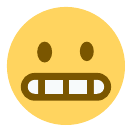}&6,031&~~~~&\emoji{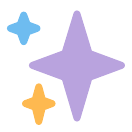}&3,072&~~~~&\emoji{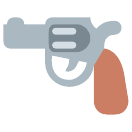}&243\\
\emoji{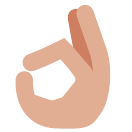}&10,104&~~~~&\emoji{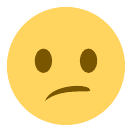}&5,849&~~~~&\emoji{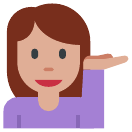}&3,088&~~~~&\emoji{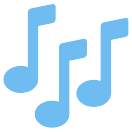}&154\\
\emoji{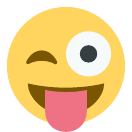}&9,546&~~~~&\emoji{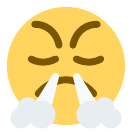}&5,624&~~~~&\emoji{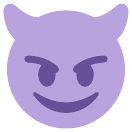}&2,969&~~~~&\emoji{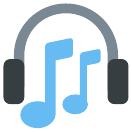}&130\\
\end{tabular}
\caption{All 64 emoji labels, and number of conversations labeled by each emoji.}
\label{tab:emoji_table}
\end{table}

\section{Dataset}

We start by describing our dataset and approaches to collecting and processing the data. Social media is a natural source of conversations, and people use emojis extensively within their posts. However, not all emojis are used to express emotion and frequency of emojis are unevenly distributed. Inspired by DeepMoji~\cite{felbo2017using}, we use 64 common emojis as labels (see Table~\ref{tab:emoji_table}), and collect a large corpus of Twitter conversations containing those emojis. Note that emojis with the difference only in skin tone are considered the same emoji.

\subsection{Data Collection}
We crawled conversation pairs consisting of an original post and a response on Twitter from 12th to 14th of August, 2017. The response to a conversation must include at least one of the 64 emoji labels. Due to the limit of Twitter streaming API, tweets are filtered on the basis of words. In our case, a tweet can be reached only if at least one of the 64 emojis is used as a word, meaning it has to be a single character separated by blank space. However, this kind of tweets is arguably cleaner, as it is often the case that this emoji is used to wrap up the whole post and clusters of repeated emojis are less likely to appear in such tweets.

For both original tweets and responses, only English tweets without multimedia contents (such as URL, image or video) are allowed, since we assume that those contents are as important as the text itself for the machine to understand the conversation. If a tweet contains less than three alphabetical words, the conversation is not included in the dataset. 

\subsection{Emoji Labeling}
Then we label responses with emojis. If there are multiple types of emoji in a response, we use the emoji with most occurrences inside the response. Among those emojis with same occurrences, we choose the least frequent one across the whole corpus, on the hypothesis that less frequent tokens better represent what the user wants to express. See Figure~\ref{fig:process_example} for example.


\begin{figure}[t]
\setlength\tabcolsep{1.8pt}
\renewcommand{\arraystretch}{1.2}
\begin{tabular}{ll}
\textbf{Before:}&@amy \bigmoji{12} miss you soooo much!!! \bigmoji{1}\\
&\bigmoji{1}\bigmoji{1}\\
\textbf{After:}&\bigmoji{12} miss you soo much! \bigmoji{1}\\
\textbf{Label:}&\bigmoji{1}\\
\end{tabular}
\caption{An artificial example illustrating preprocess procedure and choice of emoji label. Note that emoji occurrences in responses are counted before the deduplication process.}
\label{fig:process_example}
\end{figure}

\subsection{Data Preprocessing}
During preprocessing, all mentions and hashtags are removed, and punctuation\footnote{Emoticons (e.g. `:)', `(-:') are made of mostly punctuation marks. They are not examined in this paper. Common emoticons are treated as words during preprocessing.} and emojis are separated if they are adjacent to words. Words with digits are all treated as the same special token.

In some cases, users use emojis and symbols in a cluster to express emotion extensively. To normalize the data, words with more than two repeated letters, symbol strings of more than one repeated punctuation symbols or emojis are shortened, for example, `!!!!' is shortened to `!', and `yessss' to `yess'. Note that we do not reduce duplicate letters completely and convert the word to the `correct' spelling (`yes' in the example) since the length of repeated letters represents the intensity of emotion. By distinguishing `yess' from `yes', the emotional intensity is partially preserved in our dataset.

Then all symbols, emojis, and words are tokenized. Finally, we build a vocabulary of size 20K according to token frequency. Any tokens outside the vocabulary are replaced by the same special token.

We randomly split the corpus into 596,959 /32,600/32,600 conversation pairs for train /validation/test set\footnote{We will release the dataset with all tweets in its original form before preprocessing. To comply with Twitter's policy, we will include the tweet IDs in our release, and provide a script for downloading the tweets using the official API. No information of the tweet posters is collected.}. Distribution of emoji labels within the corpus is presented in Table~\ref{tab:emoji_table}.

\section{Generative Models}

In this work, our goal is to generate emotional responses to tweets with the emotion specified by an emoji label. We assembled several generative models and trained them on our dataset.

\subsection{Base: Attention-Based Sequence-to-Sequence Model}
\begin{figure}[t]
\centering
\includegraphics[width=\linewidth]{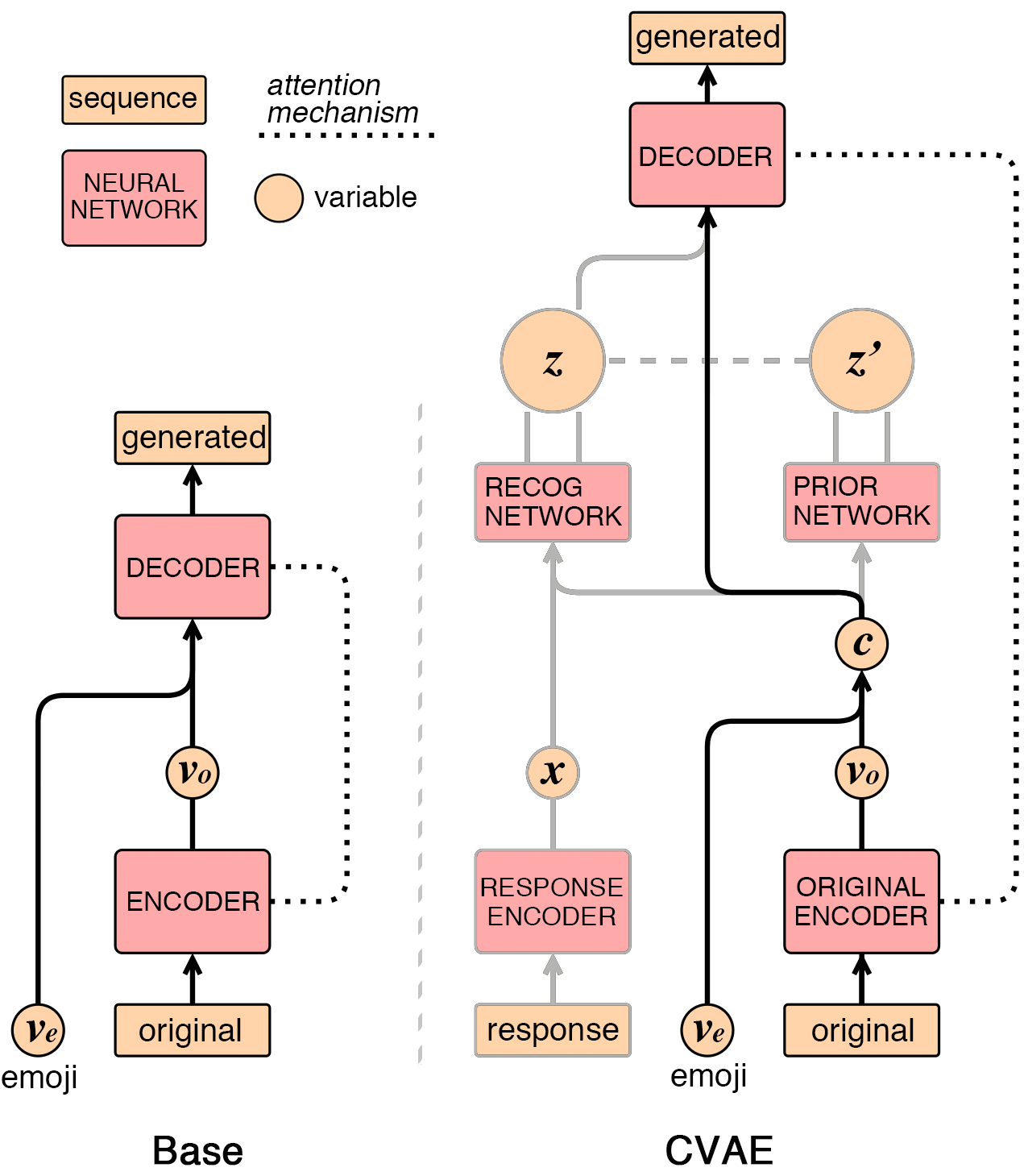}
\caption{From bottom to top is a forward pass of data during training. \textbf{Left}: the base model encodes the original tweets in $v_o$, and generates responses by decoding from the concatenation of $v_o$ and the embedded emoji, $v_e$. \textbf{Right}: In the CVAE model, all additional components (outlined in gray) can be added incrementally to the base model. A separate encoder encodes the responses in $x$. Recognition network inputs $x$ and produces the latent variable $z$ by reparameterization trick. During training, The latent variable $z$ is concatenated with $v_o$ and $v_e$ and fed to the decoder.
}
\label{fig:model}
\end{figure}

Traditional studies use deep recurrent architecture and encoder-decoder models to generate conversation responses, mapping original texts to target responses. Here we use a sequence-to-sequence (\textsc{seq2seq}) model~\cite{sutskever2014sequence} with global attention mechanism ~\cite{luong2015effective} as our base model (See Figure~\ref{fig:model}).

We use randomly initialized embedding vectors to represent each word. To specifically model the emotion, we compute the embedding vector of the emoji label the same way as word embeddings. The emoji embedding is further reduced to smaller size vector $v_e$ through a dense layer. We pass the embeddings of original tweets through a bidirectional RNN encoder of GRU cells~\cite{schuster1997bidirectional,chung2014empirical}. The encoder outputs a vector $v_o$ that represents the original tweet. Then $v_o$ and $v_e$ are concatenated and fed to a 1-layer RNN decoder of GRU cells. A response is then generated from the decoder.

\subsection{Conditional Variational Autoencoder (CVAE)}
Having similar encoder-decoder structures, \textsc{seq2seq} can be easily extended to 
a Conditional Variational Autoencoder (CVAE)~\cite{sohn2015learning}. Figure~\ref{fig:model} illustrates the model: response encoder, recognition network, and prior network are added on top of the \textsc{seq2seq} model. Response encoder has the same structure to original tweet encoder, but it has separate parameters. We use embeddings to represent Twitter responses and pass them through response encoder.

Mathematically, CVAE is trained by maximizing a variational lower bound on the conditional likelihood of $x$ given $c$, according to: 
\begin{equation}
p(x|c)=\int p(x|z,c)p(z|c)dz
\end{equation}
$z$, $c$ and $x$ are random variables. $z$ is the latent variable. In our case, the condition $c=[v_o;v_e]$, target $x$ represents the response. Decoder is used to approximate $p(x|z,c)$, denoted as $p_D(x|z,c)$. Prior network is introduced to approximate $p(z|c)$, denoted as $p_P(z|c)$. Recognition network $q_R(z|x,c)$ is introduced to approximate true posterior $p(z|x,c)$ and will be absent during generation phase. By assuming that the latent variable has a multivariate Gaussian distribution with a diagonal covariance matrix, the lower bound to $\log p(x|c)$ can then be written by:
\begin{equation} \label{eq:cvae_objective}
\begin{split}
-\mathcal{L}(\theta_D,\theta_P,\theta_R;x,c)=\text{KL}(q_R(z|x,c)||p_P(z|c)) \\
-\mathbb{E}_{q_R(z|x,c)}(\log p_D(x|z,c))
\end{split}
\end{equation}
 $\theta_D$, $\theta_P$, $\theta_R$ are parameters of those networks.
 
In recognition/prior network, we first pass the variables through an MLP to get the mean and log variance of $z$'s distribution. Then we run a reparameterization trick~\cite{kingma2013auto} to sample latent variables. During training, $z$ by the recognition network is passed to the decoder and trained to approximate $z'$ by the prior network. While during testing, the target response is absent, and $z'$ by the prior network is passed to the decoder. 

Our CVAE inherits the same attention mechanism from the base model connecting the original tweet encoder to the decoder, which makes our model deviate from previous works of CVAE on text data. Based on the attention memory as well as $c$ and $z$, a response is finally generated from the decoder. 

When handling text data, the VAE models that apply recurrent neural networks as the structure of their encoders/decoders may first learn to ignore the latent variable, and explain the data with the more easily optimized decoder. The latent variables lose its functionality, and the VAE deteriorates to a plain \textsc{seq2seq} model mathematically~\cite{bowman2015generating}. Some previous methods effectively alleviate this problem. Such methods are also important to keep a balance between the two items of the loss, namely KL loss and reconstruction loss. We use techniques of KL annealing, early stopping~\cite{bowman2015generating} and bag-of-word loss~\cite{zhao2017learning} in our models. The general loss with bag-of-word loss (see supplementary materials for details) is rewritten as:
\begin{equation} \label{eq:cvae_objective_bow}
\mathcal{L'}=\mathcal{L}+\mathcal{L}_{bow}
\end{equation}

\subsection{Reinforced CVAE}

In order to further control the emotion of our generation more explicitly, we combine policy gradient techniques on top of the CVAE above and proposed Reinforced CVAE model for our task. We first train an emoji classifier on our dataset separately and fix its parameters thereafter. The classifier is used to produce reward for the policy training. It is a skip connected model of Bidirectional GRU-RNN layers ~\cite{felbo2017using}.

During the policy training, we first get the generated response $x'$ by passing $x$ and $c$ through the CVAE, then feeding generation $x'$ to classifier and get the probability of the emoji label as reward $R$. Let $\theta$ be parameters of our network, \textsc{reinforce} algorithm~\cite{williams1992simple} is used to maximize the expected reward of generated responses:
\begin{equation} \label{eq:reinforce}
\mathcal{J}(\theta)=\mathbb{E}_{p(x|c)}(R_{\theta}(x, c))
\end{equation}
The gradient of Equation~\ref{eq:reinforce} is approximated using the likelihood ratio trick~\cite{glynn1990likelihood,williams1992simple}:
\begin{equation}
\nabla\mathcal{J}(\theta)=(R-r)\nabla\sum_t^{|x|}\log p(x_t|c,x_{1:t-1})
\end{equation}
$r$ is the baseline value to keep estimate unbiased and reduce its variance. In our case, we directly pass $x$ through emoji classifier and compute the probability of the emoji label as $r$. The model then encourages response generation that has $R>r$.

As \textsc{reinforce} objective is unrelated to response generation, it may make the generation model quickly deteriorate to some generic responses. To stabilize the training process, we propose two straightforward techniques to constrain the policy training:
\begin{enumerate}
\item Adjust rewards according to the position of the emoji label when all labels are ranked from high to low in order of the probability given by the emoji classifier. When the probability of the emoji label is of high rank among all possible emojis, we assume that the model has succeeded in emotion expression, thus there is no need to adjust parameters toward higher probability in this response. Modified policy gradient is written as:
\begin{equation} \label{eq:modified_reward}
\begin{split}
\nabla\mathcal{J}'(\theta)=\alpha(R-r)\nabla\sum_t^{|x|}\log p(x_t|c,x_{1:t-1})
\end{split}
\end{equation}
where $\alpha\in[0,1]$ is a variant coefficient. The higher $R$ ranks in all types of emoji label, the closer $\alpha$ is to 0.

\item Train Reinforced CVAE by a hybrid objective of \textsc{reinforce} and variational lower bound objective, learning towards both emotion accuracy and response appropriateness:
\begin{equation} \label{eq:hybrid_objective}
\mathbf{min}_{\theta}\mathcal{L}''=\mathcal{L}'-\lambda\mathcal{J}'
\end{equation}
$\lambda$ is a balancing coefficient, which is set to 1 in our experiments. 
\end{enumerate}

The algorithm outlining the training process of Reinforced CVAE can be found in the supplementary materials.


\section{Experimental Results and Analyses}

We conducted several experiments to finalize the hyper-parameters of our models (Table~\ref{tab:general}). During training, fully converged base \textsc{seq2seq} model is used to initialize its counterparts in CVAE models. Pretraining is vital to the success of our models since it is essentially hard for them to learn a latent variable space from total randomness.
For more details, please refer to the supplementary materials.

In this section, we first report and analyze the general results of our models, including perplexity, loss and emotion accuracy. Then we take a closer look at the generation quality as well as our models' capability of expressing emotion.

\subsection{General}
To generally evaluate the performance of our models, we use generation perplexity and top-1/top-5 emoji accuracy on the test set. Perplexity indicates how much difficulty the model is having when generating responses. We also use top-5 emoji accuracy, since the meaning of different emojis may overlap with only a subtle difference. The machine may learn that similarity and give multiple possible labels as the answer.

Note that we use the same emoji classifier for evaluation. Its accuracy (see supplementary materials) may not seem perfect, but it is the state-of-the-art emoji classifier given so many classes. Also, it's reasonable to use the same classifier in training for automated evaluation, as is in ~\cite{hu2017toward}. We can obtain meaningful results as long as the classifier is able to capture the semantic relationship between emojis~\cite{felbo2017using}.

As is shown in Table~\ref{tab:general}, CVAE significantly reduces the perplexity and increases the emoji accuracy over base model. Reinforced CVAE also adds to the emoji accuracy at the cost of a slight increase in perplexity. These results confirm that proposed methods are effective toward the generation of emotional responses.
\begin {table}[t]
\renewcommand{\arraystretch}{1.3}
\centering
\small
\begin{tabular}{l|ccc}
\hline
 & & \multicolumn{2}{c}{Emoji Accuracy} \\
\cline{3-4}
Model          & Perplexity  & Top1  & Top5\\
\hline
& \multicolumn{3}{c}{Development} \\
\hline
Base        & 127.0    & 34.2\% & 57.6\%      \\
CVAE            & 37.1     & 40.7\%    & 75.3\%      \\
Reinforced CVAE & 38.1     & 42.2\% & 76.9\%      \\
\hline
& \multicolumn{3}{c}{Test} \\
\hline
Base        & 130.6    & 33.9\% & 58.1\%      \\
CVAE            & 36.9     & 41.4\%    & 75.1\%      \\
Reinforced CVAE & 38.3     & 42.1\% & 77.3\%      \\
\hline
\end{tabular}
\caption {Generation perplexity and emoji accuracy of the three models.} \label{tab:general}
\end{table}


When converged, the KL loss is 27.0/25.5 for the CVAE/Reinforced CVAE respectively, and reconstruction loss 42.2/40.0. The models achieved a balance between the two items of loss, confirming that they have successfully learned a meaningful latent variable.

\begin{figure}[t]
\centering
\includegraphics[width=0.95\linewidth]{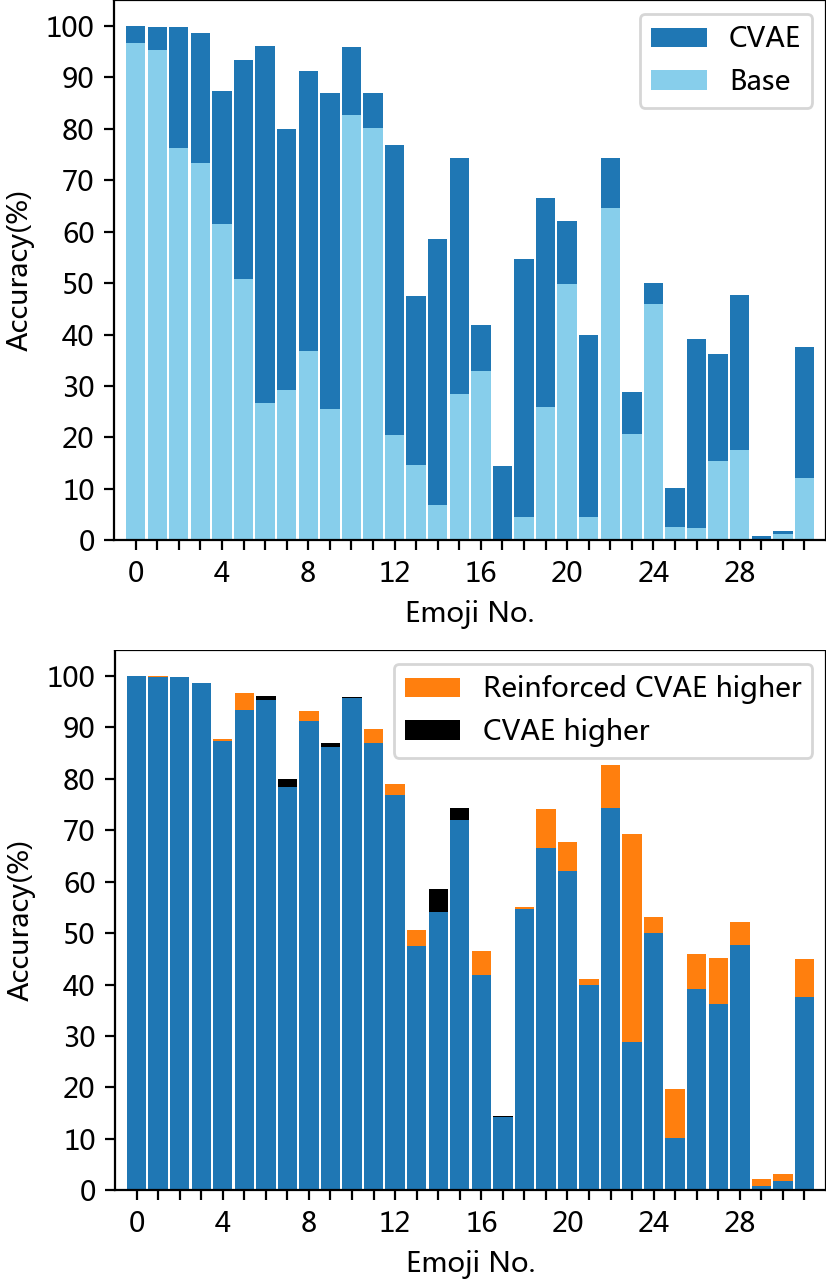}
\caption{
Top5 emoji accuracy of the first 32 emoji labels. Each bar represents an emoji and its length represents how many of all responses to the original tweets are top5 accurate. Different colors represent different models. Emojis are numbered in the order of frequencies in the dataset. No.0 is \bigmoji{0}, for instance, No.1 \bigmoji{1} and so on.\\ 
\textbf{Top:} CVAE v. Base.\\
\textbf{Bottom:} Reinforced CVAE v. CVAE. If Reinforced CVAE scores higher, the margin is marked in orange. If lower, in black.
}
\label{fig:emoji_condition}
\end{figure}

\subsection{Generation Diversity}
\textsc{Seq2seq} generates in a monotonous way, as several generic responses occur repeatedly, while the generation of CVAE models is of much more diversity. To showcase this disparity, we calculated the type-token ratios of unigrams/bigrams/trigrams in generated responses as the diversity score.

As shown in Table~\ref{tab:diversity}, results show that CVAE models beat the base models by a large margin. Diversity scores of Reinforced CVAE are reasonably compromised since it's generating more emotional responses.

\begin {table}[t]
\centering
\small
\renewcommand{\arraystretch}{1.3}
\begin{tabular}{l|ccc}
\hline
Model          & Unigram  & Bi-  & Tri-\\
\hline
\hline
Base        & 0.0061 & 0.0199 & 0.0362      \\
CVAE            & 0.0191  & 0.131  & 0.365      \\
Reinforced CVAE & 0.0160  & 0.118  & 0.337      \\
\hline
\hline
Target responses & 0.0353  & 0.370  & 0.757      \\
\hline
\end{tabular}
\caption {Type-token ratios of the generation by the three models. Scores of tokenized human-generated target responses are given for reference.}
\label{tab:diversity} 
\end{table}

\subsection{Controllability of Emotions}
There are potentially multiple types of emotion in reaction to an utterance. Our work makes it possible to generate a response to an arbitrary emotion by conditioning the generation on a specific type of emoji. In this section, we generate one response in reply to each original tweet in the dataset and condition on each emoji of the selected 64 emojis. 
We may have recorded some original tweets with different replies in the dataset, but an original tweet only need to be used once for each emoji, so we eliminate duplicate original tweets in the dataset. There are 30,299 unique original tweets in the test set.

Figure~\ref{fig:emoji_condition} shows the top-5 accuracy of each type of the first 32 emoji labels when the models generates responses from the test set conditioning on the same emoji. The results show that CVAE models increase the accuracy over every type of emoji label. Reinforced CVAE model sees a bigger increase on the less common emojis, confirming the effect of the emoji-specified policy training.


\begin {table}[t]
\small
\centering
\renewcommand{\arraystretch}{1.3}
\begin{tabular}{l|l|ccc}
\hline
Setting & Model v. Base & Win    & Lose   & Tie\\
\hline
\hline
reply & CVAE             & 42.4\% & 43.0\% & 14.6\%    \\
reply & Reinforced CVAE  & 40.6\% & 39.6\% & 19.8\%    \\
\hline
emoji & CVAE             & 48.4\% & 26.2\% & 25.4\%    \\
emoji & Reinforced CVAE  & 50.0\% & 19.6\% & 30.4\%    \\
\hline
\end{tabular}
\caption {Results of human evaluation. Tests are conducted pairwise between CVAE models and the base model.} \label{tab:human} 
\end{table}

\begin{table*}[!t]
\centering
\renewcommand{\arraystretch}{1.5}
\small
	\begin{tabular}{|>{\bfseries}p{.132\linewidth}|*3{p{.252\linewidth}|}}
		\hline
		Content & \multicolumn{3}{p{.756\linewidth}|}{sorry guys , was gunna stream tonight but i 'm still feeling like crap and my voice disappeared . i will make it up to you}\\ \hline
		Target Emotion & \emoji{5} & \emoji{17} & \emoji{20} \\ \hline
		Base & i 'm sorry you 're going to be missed it & i 'm sorry for your loss & i 'm sorry you 're going to be able to get it \\ \hline
		CVAE & hope you are okay hun ! & hi jason , i 'll be praying for you & im sorry u better suck u off \\ \hline
		Reinforced CVAE & hope you 're feeling it & hope you had a speedy recovery man ! hope you feel better soon , please get well soon & dude i 'm so sorry for that i wanna hear it and i 'm sorry i can 't go to canada with you but i wanna be away from canada \\ \hline
		\hline
		Content & \multicolumn{3}{p{.756\linewidth}|}{add me in there my bro \emoji{17}}\\ \hline
		Target Emotion & \emoji{15} & \emoji{40} & \emoji{59} \\ \hline
		Base & i 'm not sure you 'll be there & i 'm here for you & i 'm not ready for you \\ \hline
		CVAE & you know , you need to tell me in your hometown ! & you will be fine bro , i 'll be in the gym for you & i can 't wait \emoji{21}\\ \hline
		Reinforced CVAE & you might have to get me hip hop off . & good luck bro ! this is about to be healthy & i 'm still undecided and i 'm still waiting \\ \hline
		\hline
		Content & \multicolumn{3}{p{.756\linewidth}|}{don 't tell me match of the day is delayed because of this shit}\\ \hline
		Target Emotion & \emoji{0} & \emoji{4} & \emoji{37} \\ \hline
		Base & i 'm not even a fan of the game &  i 'm not sure if you ever have any chance to talk to someone else & i 'm sorry i 'm not doubting you \\ \hline
		CVAE & you can 't do it bc you 're in my mentions & see now a good point & hiya , unfortunately , it 's not \\ \hline
		Reinforced CVAE & oh my god i 'm saying this as long as i remember my twitter & fab mate , you 'll enjoy the game and you 'll get a win & it 's the worst \\ \hline
		\hline
		Content & \multicolumn{3}{p{.756\linewidth}|}{g i needed that laugh lmfaoo}\\ \hline
		Target Emotion & \emoji{2} & \emoji{19} & \emoji{37} \\ \hline
		Base & i 'm glad you enjoyed it &  i 'm not gonna lie & i 'm sorry i 'm not laughing \\ \hline
		CVAE & good ! have a good time & i don 't plan on that & me too . but it 's a lot of me . \\ \hline
		Reinforced CVAE & thank you for your tweet , you didn 't know how much i guess & that 's a bad idea , u gotta hit me up on my phone & i feel bad at this and i hope you can make a joke \\ \hline
	\end{tabular}
\caption{Some examples from our generated emotional responses. Context is the original tweet, and target emotion is specified by the emoji. Following are the responses generated by each of the three models based on the context and the target emotion.}
\label{fig:examples}
\end{table*}


\subsection{Human Evaluation}
We employed crowdsourced judges to evaluate a random sample of 100 items (Table ~\ref{tab:human}), each being assigned to 5 judges on the Amazon Mechanical Turk. We present judges original tweets and generated responses. In the first setting of human evaluation, judges are asked to decide which one of the two generated responses better reply the original tweet. In the second setting, the emoji label is presented with the item discription, and judges are asked to pick one of the two generated responses that they decide better fits this emoji. (These two settings of evaluation are conducted separately so that it will not affect judges' verdicts.) Order of two generated responses under one item is permuted. Ties are permitted for answers. We batch five items as one assignment and insert an item with two identical outputs as the sanity check. Anyone who failed to choose ``tie'' for that item is considered as a careless judge and is therefore rejected from our test.

We then conducted a simplified Turing test. Each item we present judges an original tweet, its reply by a human, and its response generated from Reinforced CVAE model. We ask judges to decide which of the two given responses is written by a human. Other parts of the setting are similar to above-mentioned tests. It turned out 18\% of the test subjects mistakenly chose machine-generated responses as human written, and 27\% stated that they were not able to distinguish between the two responses.

In regard of the inter-rater agreement, there are four cases. The ideal situation is that all five judges choose the same answer for a item, and in the worst-case scenario, at most two judges choose the same answer. In light of this, we have counted that 32\%/33\%/31\%/5\% of all items have 5/4/3/2 judges in agreement, showing that our experiment has a reasonably reliable inter-rater agreement.

\subsection{Case Study}

We sampled some generated responses from all three models, and list them in Figure~\ref{fig:examples}. 
Given an original tweet, we would like to generate responses with three different target emotions.

\textsc{Seq2seq} only chooses to generate  most frequent expressions, forming a predictable pattern for its generation (See how every sampled response by the base model starts with ``I'm''). On the contrary, generation from the CVAE model is diverse, which is in line with previous quantitative analysis. However, the generated responses are sometimes too diversified and unlikely to reply to the original tweet.


Reinforced CVAE somtetimes tends to generate a lengthy response by stacking up sentences (See the responses to the first tweet when conditioning on the `folded hands' emoji and the `sad face' emoji). It learns to break the length limit of sequence generation during hybrid training, since the variational lower bound objective is competing with \textsc{reinforce} objective. The situation would be more serious is $\lambda$ in Equation~\ref{eq:hybrid_objective} is set higher. However, this phenomenon does not impair the fluency of generated sentences, as can be seen in Figure~\ref{fig:examples}.

\section{Conclusion and Future Work}
In this paper, we investigate the possibility of using naturally annotated emoji-rich Twitter data for emotional response generation. More specifically, we collected more than half a million Twitter conversations with emoji in the response and assumed that the fine-grained emoji label\ chosen by the user expresses the emotion of the tweet. We applied several state-of-the-art neural models to learn a generation system that is capable of giving a response with an arbitrarily designated emotion. We performed automatic and human evaluations to understand the quality of generated responses. We trained a large scale emoji classifier and ran the classifier on the generated responses to evaluate the emotion accuracy of the generated response. We performed an Amazon Mechanical Turk experiment, by which we compared our models with a baseline sequence-to-sequence model on metrics of relevance and emotion. Experimentally, it is shown that our model is capable of generating high-quality emotional responses, without the need of laborious human annotations.
Our work is a crucial step towards building intelligent dialog agents. We are also looking forward to transferring the idea of naturally-labeled emojis to task-oriented dialog and multi-turn dialog generation problems. Due to the nature of social media text, some emotions, such as fear and disgust, are underrepresented in the dataset, and the distribution of emojis is unbalanced to some extent. 
We will keep accumulating data and increase the ratio of underrepresented emojis, and advance toward more sophisticated abstractive generation methods.




\bibliography{acl2018.bib}

\begin{thebibliography}{28}
\expandafter\ifx\csname natexlab\endcsname\relax\def\natexlab#1{#1}\fi

\bibitem[{Bowman et~al.(2015)Bowman, Vilnis, Vinyals, Dai, J{\'{o}}zefowicz,
  and Bengio}]{bowman2015generating}
Samuel~R. Bowman, Luke Vilnis, Oriol Vinyals, Andrew~M. Dai, Rafal
  J{\'{o}}zefowicz, and Samy Bengio. 2015.
\newblock Generating sentences from a continuous space.
\newblock \emph{CONLL}.

\bibitem[{Chung et~al.(2014)Chung, G{\"{u}}l{\c{c}}ehre, Cho, and
  Bengio}]{chung2014empirical}
Junyoung Chung, {\c{C}}aglar G{\"{u}}l{\c{c}}ehre, KyungHyun Cho, and Yoshua
  Bengio. 2014.
\newblock Empirical evaluation of gated recurrent neural networks on sequence
  modeling.
\newblock \emph{NIPS 2014 Deep Learning and Representation Learning Workshop}.

\bibitem[{Eisner et~al.(2016)Eisner, Rockt{\"a}schel, Augenstein,
  Bo{\v{s}}njak, and Riedel}]{eisner2016emoji2vec}
Ben Eisner, Tim Rockt{\"a}schel, Isabelle Augenstein, Matko Bo{\v{s}}njak, and
  Sebastian Riedel. 2016.
\newblock emoji2vec: Learning emoji representations from their description.
\newblock \emph{SocialNLP at EMNLP}.

\bibitem[{Felbo et~al.(2017)Felbo, Mislove, S{\o}gaard, Rahwan, and
  Lehmann}]{felbo2017using}
Bjarke Felbo, Alan Mislove, Anders S{\o}gaard, Iyad Rahwan, and Sune Lehmann.
  2017.
\newblock Using millions of emoji occurrences to learn any-domain
  representations for detecting sentiment, emotion and sarcasm.
\newblock \emph{EMNLP}.

\bibitem[{Glorot and Bengio(2010)}]{glorot2010understanding}
Xavier Glorot and Yoshua Bengio. 2010.
\newblock Understanding the difficulty of training deep feedforward neural
  networks.
\newblock In \emph{International Conference on Artificial Intelligence and
  Statistics}, pages 249--256.

\bibitem[{Glynn(1990)}]{glynn1990likelihood}
Peter~W Glynn. 1990.
\newblock Likelihood ratio gradient estimation for stochastic systems.
\newblock \emph{Communications of the ACM}, 33(10):75--84.

\bibitem[{Go et~al.(2016)Go, Bhayani, and Huang}]{go2016sentiment140}
Alec Go, Richa Bhayani, and Lei Huang. 2016.
\newblock Sentiment140.
\newblock \url{http://help.sentiment140.com/}.

\bibitem[{Hochreiter and Schmidhuber(1997)}]{hochreiter1997long}
Sepp Hochreiter and J{\"u}rgen Schmidhuber. 1997.
\newblock Long short-term memory.
\newblock \emph{Neural computation}, 9(8):1735--1780.

\bibitem[{Hu et~al.(2017)Hu, Yang, Liang, Salakhutdinov, and
  Xing}]{hu2017toward}
Zhiting Hu, Zichao Yang, Xiaodan Liang, Ruslan Salakhutdinov, and Eric~P Xing.
  2017.
\newblock Toward controlled generation of text.
\newblock In \emph{International Conference on Machine Learning}, pages
  1587--1596.

\bibitem[{Huang et~al.(2017)Huang, Labetoulle, Huang, Chen, Chen, Srivastava,
  and Ku}]{huang2017moodswipe}
Chieh-Yang Huang, Tristan Labetoulle, Ting-Hao~Kenneth Huang, Yi-Pei Chen,
  Hung-Chen Chen, Vallari Srivastava, and Lun-Wei Ku. 2017.
\newblock Moodswipe: A soft keyboard that suggests messages based on
  user-specified emotions.
\newblock \emph{EMNLP Demo}.

\bibitem[{Kingma and Welling(2013)}]{kingma2013auto}
Diederik~P Kingma and Max Welling. 2013.
\newblock Auto-encoding variational bayes.
\newblock \emph{ICLR}.

\bibitem[{Larsen et~al.(2015)Larsen, S{\o}nderby, Larochelle, and
  Winther}]{larsen2015autoencoding}
Anders Boesen~Lindbo Larsen, S{\o}ren~Kaae S{\o}nderby, Hugo Larochelle, and
  Ole Winther. 2015.
\newblock Autoencoding beyond pixels using a learned similarity metric.
\newblock \emph{ICML}.

\bibitem[{Li et~al.(2016)Li, Monroe, Ritter, and Jurafsky}]{li2016deep}
Jiwei Li, Will Monroe, Alan Ritter, and Dan Jurafsky. 2016.
\newblock Deep reinforcement learning for dialogue generation.
\newblock \emph{EMNLP}.

\bibitem[{Li et~al.(2017{\natexlab{a}})Li, Monroe, Shi, Ritter, and
  Jurafsky}]{li2017adversarial}
Jiwei Li, Will Monroe, Tianlin Shi, Alan Ritter, and Dan Jurafsky.
  2017{\natexlab{a}}.
\newblock Adversarial learning for neural dialogue generation.
\newblock \emph{EMNLP}.

\bibitem[{Li et~al.(2017{\natexlab{b}})Li, Su, Shen, Li, Cao, and
  Niu}]{li2017dailydialog}
Yanran Li, Hui Su, Xiaoyu Shen, Wenjie Li, Ziqiang Cao, and Shuzi Niu.
  2017{\natexlab{b}}.
\newblock Dailydialog: A manually labelled multi-turn dialogue dataset.
\newblock \emph{IJCNLP}.

\bibitem[{Liu(2012)}]{liu2012sentiment}
Bing Liu. 2012.
\newblock Sentiment analysis and opinion mining.
\newblock \emph{Synthesis lectures on human language technologies},
  5(1):1--167.

\bibitem[{Luong et~al.(2015)Luong, Pham, and Manning}]{luong2015effective}
Minh-Thang Luong, Hieu Pham, and Christopher~D Manning. 2015.
\newblock Effective approaches to attention-based neural machine translation.
\newblock \emph{EMNLP}.

\bibitem[{Maas et~al.(2011)Maas, Daly, Pham, Huang, Ng, and
  Potts}]{maas2011learning}
Andrew~L Maas, Raymond~E Daly, Peter~T Pham, Dan Huang, Andrew~Y Ng, and
  Christopher Potts. 2011.
\newblock Learning word vectors for sentiment analysis.
\newblock In \emph{Proceedings of the 49th Annual Meeting of the Association
  for Computational Linguistics: Human Language Technologies-Volume 1}, pages
  142--150. Association for Computational Linguistics.

\bibitem[{Mikolov et~al.(2013)Mikolov, Chen, Corrado, and
  Dean}]{mikolov2013efficient}
Tomas Mikolov, Kai Chen, Greg Corrado, and Jeffrey Dean. 2013.
\newblock Efficient estimation of word representations in vector space.
\newblock \emph{ICLR}.

\bibitem[{Pang et~al.(2002)Pang, Lee, and Vaithyanathan}]{pang2002thumbs}
Bo~Pang, Lillian Lee, and Shivakumar Vaithyanathan. 2002.
\newblock Thumbs up?: sentiment classification using machine learning
  techniques.
\newblock In \emph{Proceedings of the ACL-02 conference on Empirical methods in
  natural language processing-Volume 10}, pages 79--86. Association for
  Computational Linguistics.

\bibitem[{Pang et~al.(2008)Pang, Lee et~al.}]{pang2008opinion}
Bo~Pang, Lillian Lee, et~al. 2008.
\newblock Opinion mining and sentiment analysis.
\newblock \emph{Foundations and Trends{\textregistered} in Information
  Retrieval}, 2(1--2):1--135.

\bibitem[{Schuster and Paliwal(1997)}]{schuster1997bidirectional}
Mike Schuster and Kuldip~K Paliwal. 1997.
\newblock Bidirectional recurrent neural networks.
\newblock \emph{IEEE Transactions on Signal Processing}, 45(11):2673--2681.

\bibitem[{Socher et~al.(2013)Socher, Perelygin, Wu, Chuang, Manning, Ng, and
  Potts}]{socher2013recursive}
Richard Socher, Alex Perelygin, Jean Wu, Jason Chuang, Christopher~D Manning,
  Andrew Ng, and Christopher Potts. 2013.
\newblock Recursive deep models for semantic compositionality over a sentiment
  treebank.
\newblock In \emph{Proceedings of the 2013 conference on empirical methods in
  natural language processing}, pages 1631--1642.

\bibitem[{Sohn et~al.(2015)Sohn, Lee, and Yan}]{sohn2015learning}
Kihyuk Sohn, Honglak Lee, and Xinchen Yan. 2015.
\newblock Learning structured output representation using deep conditional
  generative models.
\newblock In \emph{Advances in Neural Information Processing Systems}, pages
  3483--3491.

\bibitem[{Sutskever et~al.(2014)Sutskever, Vinyals, and
  Le}]{sutskever2014sequence}
Ilya Sutskever, Oriol Vinyals, and Quoc~V Le. 2014.
\newblock Sequence to sequence learning with neural networks.
\newblock In \emph{Advances in neural information processing systems}, pages
  3104--3112.

\bibitem[{Williams(1992)}]{williams1992simple}
Ronald~J Williams. 1992.
\newblock Simple statistical gradient-following algorithms for connectionist
  reinforcement learning.
\newblock \emph{Machine learning}, 8(3-4):229--256.

\bibitem[{Xie et~al.(2016)Xie, Liu, Yan, and Sun}]{xie2016neural}
Ruobing Xie, Zhiyuan Liu, Rui Yan, and Maosong Sun. 2016.
\newblock Neural emoji recommendation in dialogue systems.
\newblock \emph{arXiv preprint arXiv:1612.04609}.

\bibitem[{Zhao et~al.(2017)Zhao, Zhao, and Eskenazi}]{zhao2017learning}
Tiancheng Zhao, Ran Zhao, and Maxine Eskenazi. 2017.
\newblock Learning discourse-level diversity for neural dialog models using
  conditional variational autoencoders.
\newblock \emph{ACL}.

\end{thebibliography}
\bibliographystyle{acl_natbib.bst}

\appendix

\section{Supplementary Materials}

\subsection{Bag-of-Word Loss}
In the idea of BoW loss, $x$ can be decomposed into $x_o$ of word order and $x_{bow}$ of words without order. By assuming that $x_o$ and $x_{bow}$ are conditionally independent, $p(x|z,c)=p(x_o|z,c)p(x_{bow}|z,c)$. Given $z$ and $c$, $p(x_{bow}|z,c)$ is the product over probability of every token in the text:
\begin{equation}
\begin{split}
p(x_{bow}|z,c)=\prod_{t=1}^{|x|}p(x_t|z,c) \\
=\prod_{t=1}^{|x|}\text{softmax}(f(x_t,z,c))
\end{split}
\end{equation}
Function $f$ first maps $z$, $c$ to space $\mathbb{R}^V$, where $V$ is the vocabulary size, and then chose the element corresponding to token $x_t$ as its logit.

Now the modified objective is written by:
\begin{equation} \label{eq:revised_cvae_objective}
\begin{split}
\mathcal{L}'(\theta_D,\theta_P,\theta_R;x,c)=\mathcal{L}(\theta_D,\theta_P,\theta_R;x,c) \\
+\mathbb{E}_{q_R(z|x,c)}(\log p(x_{bow}|z,c))
\end{split}
\end{equation}
Finally, CVAE is trained by minimizing $\mathcal{L}'$.

\begin{figure*}[!t]
\centering
\includegraphics[width=0.8\linewidth]{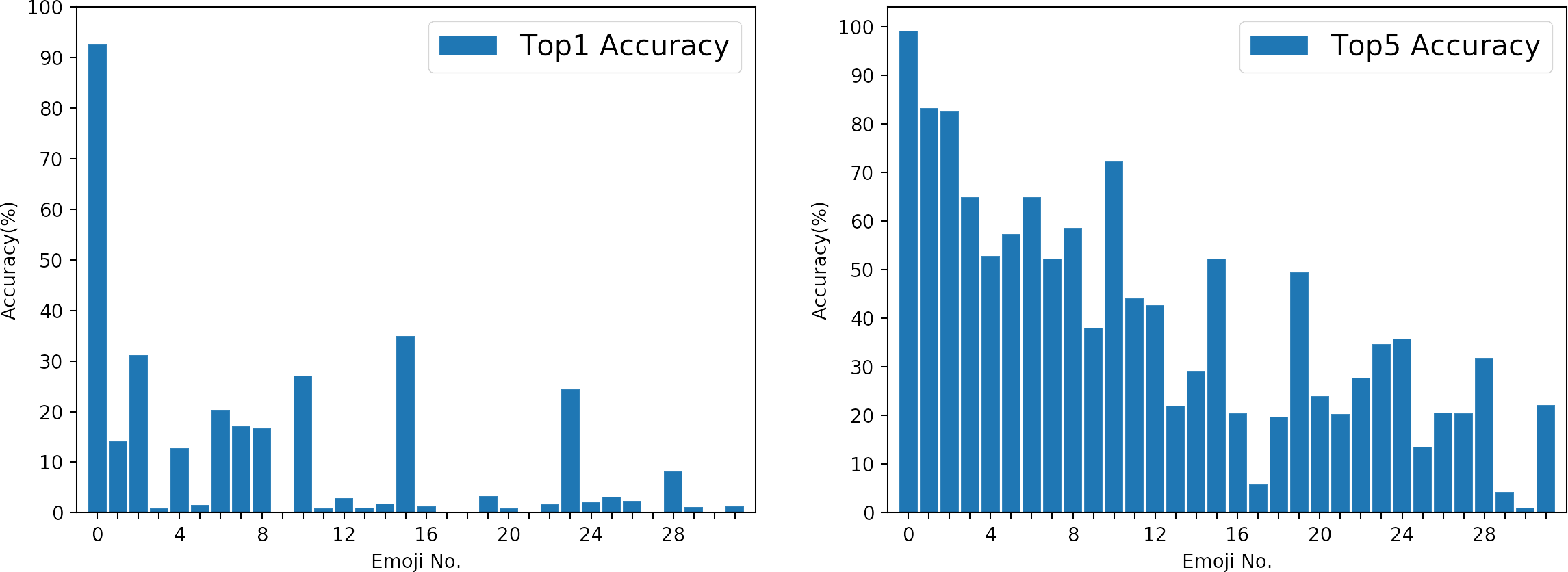}
\caption{Top-1 and top-5 accuracy of emoji classifier by each emoji label on test set.}
\label{fig:classifier_accuracy}
\end{figure*}

\subsection{Emoji Classifier}
The emoji classifier is a skip connected model of Bidirectional GRU-RNN layers and has the same structure as the classifier in~\cite{felbo2017using}. This separate neural network uses the same set of hyper-parameters (embedding size, hidden state size, etc.) as in the generation models described below. We train it on our train set by mapping response Tweets to their emoji label, with a dropout rate of 0.2 and an Adam optimizer of a 1e-3 learning rate with gradient clipped to 5. RNN layers and word embeddings in the classifier have a dimension of 128. All weights of dense layers are initialized by Glorot uniform initializer~\cite{glorot2010understanding} and word embeddings are initialized by sampling from the uniform distribution [-4e-3, 4e-3].

The classifier gives the probability of all 64 emoji labels. For 32.1\% responses in the test set, the probability of the emoji label ranks highest of all emoji labels. In 57.8\% of cases, the probability of emoji label is among the five highest. We refer to the two figures as \textit{top-1} and \textit{top-5 accuracy}. Figure~\ref{fig:classifier_accuracy} shows the top-1 and top-5 accuracy of the 32 most frequent emoji labels. Accuracy for less common emojis may be low since they are underrepresented in the dataset.

\subsection{Training Process of the Reinforced CVAE}

\begin{algorithm}[t] 
\SetAlgoLined
\SetKwInOut{Input}{input}
\SetKwInOut{Output}{output}
\SetKwProg{Procedure}{procedure}{}{}
\Input{Total training step $N$, Training batches, $\lambda$}
 Pretrain CVAE by minimizing Eq.~\ref{eq:revised_cvae_objective}\;
 $i=0$\;
 \While{$i<N$}{
  Get next batch B and target responses T in B\;
  \Procedure{Forward pass B through CVAE}{
   get generation G\;
   get probability $P$ of all words in G\;
   get variational lower bound objective $\mathcal{L}'$\;
  }
  Compute $R$, $\alpha$ by emoji classifier using G\;
  Compute $r$ by emoji classifier using T\;
  $\mathcal{J}'=\alpha(R-r)\sum\log P$\;
  $\mathcal{L}''=\mathcal{L}'-\lambda\mathcal{J}'$\;
  Conduct gradient descent on CVAE using $\mathcal{L}''$\;
  $i++$\;
 }
 \caption{Training of the Reinforced CVAE.}
 \label{alg:1}
\end{algorithm}

Algorithm~\ref{alg:1} outlines the training process of the Reinforced CVAE. The first step of pretraining is described in the next section. For every training batch, we first compute the variational objective $\mathcal{L}'$ and obtain the generated text. Then we compute the policy gradient $\mathcal{J}'$ from the word probability in the previously generated text and the rewards determined by the emoji classifier. Finally, we conduct gradient descent on the CVAE components using the hybrid objective $\mathcal{L}''$ that is comprised of $\mathcal{L}'$ and $\mathcal{J}'$.

\subsection{Experiment Setting}
\paragraph{Hyper-parameters}
For the hyper-parameters of the base model and CVAE models, we use word embeddings of 128 dimensions and RNN layers of 128 hidden units for all encoders and decoders. The size of emojis' embeddings is contracted to 12 through a dense layer of $tanh$ non-linearity. We set the size of latent variables to 268. MLPs in recognition/prior network are 3 layered with $tanh$ non-linearity. All other training settings are the same as the emoji classifier.


For Reinforced CVAE\footnote{We will release the source code for \textsc{mojitalk} and pre-trained models on Github.com.}, $\lambda$ in hybrid objective (Eq.6 of the paper) is set to 1, and $\alpha$ in Eq.5 of the paper is empirically given by:
\begin{equation}
\begin{split}
\alpha_{x',e}=\left\{
            \begin{array}{ll}
                  0,\\
                  0.5,\\
                  1,
            \end{array}
           \right.
           \begin{array}{ll}
                  R \textit{ ranks 1 in all labels}\\
                  R \textit{ ranks 2 to 5 in all labels}\\
                  \textit{otherwise}
            \end{array}
\end{split}
\end{equation}
where reward $R$ is the probability of emoji label $e$ computed by the classifier, and $x'$ is the generated response.

\paragraph{Training Setting}
We use fully converged base \textsc{seq2seq} model to initialize its counterparts in CVAE models. When training the Reinforced CVAE with emoji classifier, instead of using hybrid loss function from the beginning, we introduce the policy loss only after 2 epochs of training.

For our final models, we use bow loss along with KL annealing to 0.5 at the end of the 6th epoch. Note that KL weight does not anneal to 1 at last, meaning that our models do not strictly follow the objective of CVAE (Equation~\ref{eq:revised_cvae_objective}). However, lower KL weight gives the model more freedom to generate text. We can view this technique as early stopping~\cite{bowman2015generating}, finding a better result before model converges on the original objective.

\paragraph{Generation}
To exploit the randomness of the latent variable, during generation, we sample the result of CVAE models 5 times and choose the generated response with the highest probability of designated emoji label as the final generation.


\end{document}